\def\year{2020}\relax
\documentclass[letterpaper]{article} 
\usepackage{aaai20}  
\usepackage{times}  
\usepackage{helvet} 
\usepackage{courier}  
\usepackage[hyphens]{url}  
\usepackage{graphicx} 

\usepackage{latexsym}

\usepackage{amsmath}
\usepackage{amsfonts}
\usepackage{subfigure}
\usepackage{multirow}
\usepackage{booktabs}
\usepackage{bm}
\usepackage[normalem]{ulem}

\def\UrlFont{\rm}  
\usepackage{graphicx}  
\title{CopyMTL: Copy Mechanism for Joint Extraction of Entities and Relations with Multi-Task Learning}
\author{Daojian Zeng\thanks{Corresponding authors, equal contribution}\textsuperscript{\rm $\S$}, Ranran Haoran Zhang\footnotemark[1]\textsuperscript{\rm $\dagger$} , \Large Qianying Liu\textsuperscript{\rm $\ddagger$}\\ 
	\textsuperscript{\rm $\S$}Changsha University of Science \& Technology, Changsha, 410114, China\\ 
	\textsuperscript{\rm $\dagger$}University of Illinois at Urbana-Champaign, Illinois, 61820, USA\\ 
	\textsuperscript{\rm $\ddagger$}Kyoto University, Kyoto, 606-8501, Japan\\
	\url{zengdj916@163.com}, \ \url{haoranz6@illinois.edu}, \ \url{ying@nlp.ist.i.kyoto-u.ac.jp}
}

\begin{document}

\maketitle

\begin{abstract}
Joint extraction of entities and relations has received significant attention due to its potential of providing higher performance for both tasks. Among existing methods, 
CopyRE is effective and novel, which uses a sequence-to-sequence framework and copy mechanism to directly generate the relation triplets.
However, it suffers from two fatal problems. The model is extremely weak at differing the head and tail entity, resulting in inaccurate entity extraction. It also cannot predict multi-token entities (e.g. \textit{Steven Jobs}). To address these problems, we give a detailed analysis of the reasons behind the inaccurate entity extraction problem, and then propose a simple but extremely effective model structure to solve this problem. In addition, we propose a multi-task learning framework equipped with copy mechanism, called CopyMTL, to allow the model to predict multi-token entities. Experiments reveal the problems of CopyRE and show that our model achieves significant improvement over the current state-of-the-art method by 9\% in NYT and 16\% in WebNLG (F1 score). Our code is available at \url{https://github.com/WindChimeRan/CopyMTL}

  \end{abstract}

  \section{Introduction}
  \label{sec:intro}


  As a key technology for automatic Knowledge Graph (KG) construction, relation extraction has received widespread attention in recent years. Relation extraction aims to automatically learn triplets \textit{(relation, head, tail)} from the unstructured text without human intervention.

  Early studies use pipeline models \cite{pipeline1,pipeline2}, where they cast the relation extraction problem into two separate tasks, i.e. Named Entity Recognition (NER) to extract the entities and Relation Classification. 
  They first recognize the entities and then predict the relations between entities.
  However, pipeline models suffer from obvious drawbacks \cite{linear_programming}. Each component limits the performance because of the error cascading effect and there is no chance for the model to correct mistakes. In addition, such pipeline models cannot capture the explicit relation between the two subtasks \cite{interaction}, where joint models can benefit from such interdependencies.
  
  Recent studies on joint models of entity and relation extraction have three major research lines: Table Filling, Tagging, and Sequence-to-Sequence (Seq2Seq). Among these approaches, the table filling method \cite{table_filling,crf_norm} requires the model to enumerate over all possible entity pairs, which leads to a heavy computational burden. 
  The tagging method \cite{novel_tagging} suffers from the overlapping relation problem that the model cannot assign different relation tags to one token. To solve this problem, the followers \cite{rlre,tag2table} run tagging on a sentence for multiple turns, which is akin to the table filling method together with the heavy computational burden. 
  Relatively speaking, the Seq2Seq method is neither plagued with overlapping relations nor with excessive computations.
  Seq2Seq model receives the unstructured text as input and directly decodes the entity-relation triples as a sequential output. 
  This concise approach also matches with the human annotation process, that the annotators first read the sentences, understand the meaning and then point out the entity-relation pairs sequentially.
  
   Currently, CopyRE \cite{CopyRE} is the most powerful Seq2Seq based joint extraction method which expands a Seq2Seq framework with copy mechanism in the decoder. The copying mechanism allows the model to avoid the out-of-vocabulary (OOV) problem. Despite their promising result, the model still suffers from two major drawbacks. 
  \begin{figure}[t]
  \centering
  \includegraphics[width=0.47\textwidth]{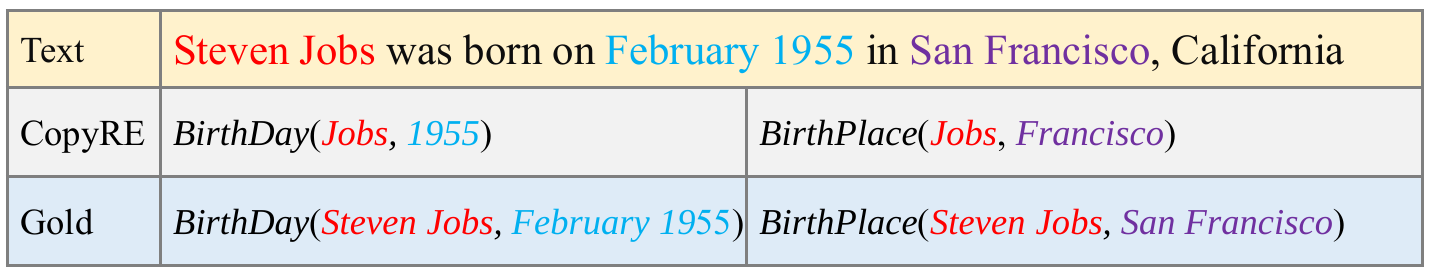}
  \caption{CopyRE predicts the entity pointer refers to the word position in the source sentence. The colored tokens show the limitation of CopyRE which cannot predict multiple tokens.}
  \label{intro}
  \end{figure}

First, the entity copying in CopyRE is unstable and it depends on an unnatural mask to differ the head ($h$) and tail ($t$) entities. Experimental results show that CopyRE nearly randomly predicts the head-tail order of the two entities. The model also needs an unnatural mask that masks the probability of $h$ while predicting $t$. Without this mask, when predicting $t$, the model would choose the same token as $h$, and the accuracy drops to zero. 
After analysis, we prove that CopyRE actually uses the same distribution to model $h$ and $t$, chooses the highest probability as $h$, and the second-highest would be chosen as $t$ after masking the highest probability, so without this mask, it cannot differ $h$ and $t$. Modeling the $h$ and $t$ distribution in such manner can cause various problems, the model not only is extremely weak at differing $h$ and $t$, but also cannot get information about $h$ while predicting $t$.
  
   Second, CopyRE cannot extract entities that have multiple tokens. The copy-based decoder always points to the last token of any entities, which limits the applicability of the model.
   For example, in Fig. \ref{intro} we show that CopyRE only predicts \textit{``Jobs"} rather than the whole entity \textit{``Steven Jobs"} when the entity has two tokens. In real-word scene multi-token entities are common, so this can greatly drag the model performance.

  To address these two problems mentioned above, we propose CopyMTL, which is a multi-task learning based model with a new architecture for entity copying. We first provide a detailed analysis of why CopyRE is unstable during copying and propose a new model architecture to improve the shortcomings. Our new model architecture merely adds one more non-linear fully connection layer so that the model predicts separate distributions for the head and tail entity, and the tail prediction receives information from the head prediction. This architecture no longer needs the unnatural mask and increases the accuracy of entity copying, resulting in the overall improvements over the state-of-the-art model.
  
  Then we propose a multi-task learning based Seq2Seq model to predict multi-token entities. A sequence labeling layer is added at the encoding stage to assist the entity recognition process. 
  We use multi-task learning of NER to predict the start token of each entity while the decoder points at the last token while decoding. During training, we optimize the multi-task loss function jointly.
  
  In conclusion, the contribution of this work is as follow:
  
  1. We analyze the reasons for the unstable performance of entity copying in CopyRE and propose a simple but effective architecture to address this problem. 
  
  2. We propose a multi-task framework to enhance the capability of handling with multi-token entities.
  
  3. Experimental results show that our model achieves state-of-the-art results and outperforms previous approaches by a large margin.

  \section{Background}
  \label{background}    
      
\begin{figure*}[t]
          \centering
          \includegraphics[width=0.6\textwidth]{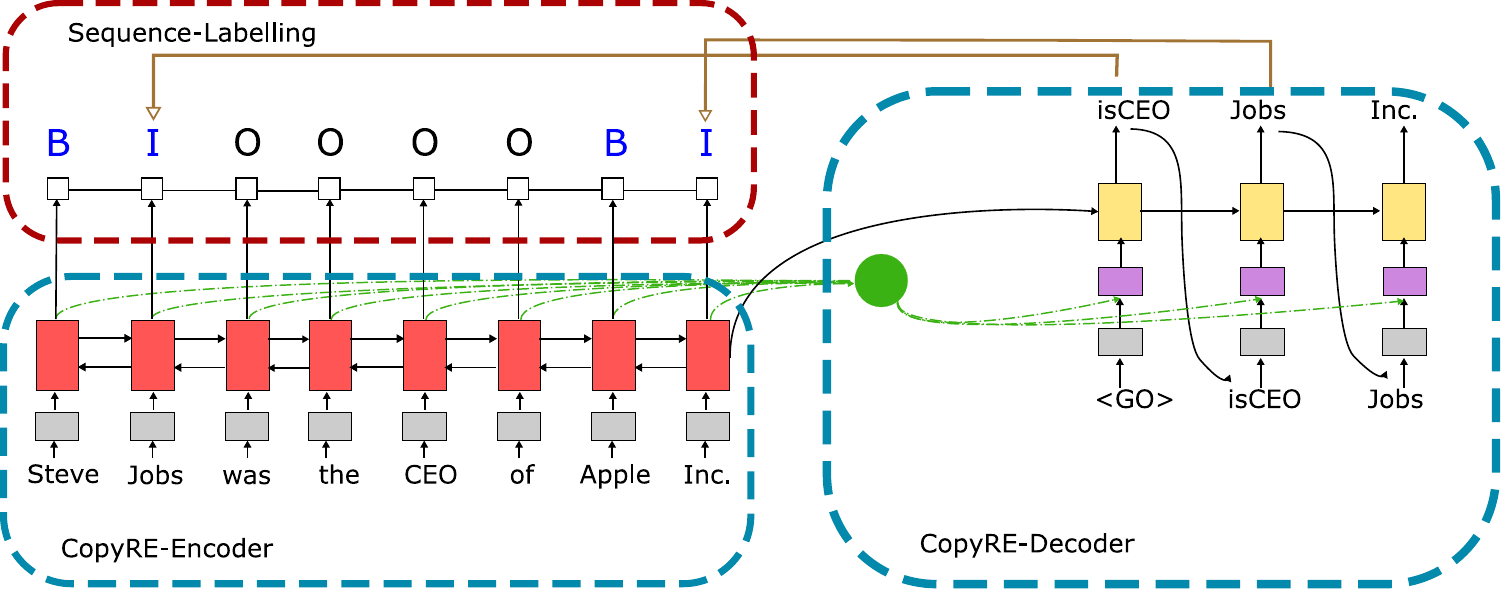}
          \caption{The overview of CopyMTL model for joint extraction of relation and entity. The CopyRE model does not contain the CopyMTL-Tagging part, i.e., the sequence-labeling part in the figure.}
          \label{fig:copymt}
      \end{figure*}

      In this section, we first introduce the CopyRE model that is based on the Seq2Seq framework. 
      Then, we give a detailed description of the two existing problems.
      As shown in Fig. \ref{fig:copymt}, CopyRE consists two parts: an encoder and a decoder.
      Given a sentence $s=\{x_1,x_2...x_n\}$, the encoder transforms the input $s$ into a vector representation. 
      The decoder predicts the relation-entity triplets $(r, h, t)$ each three time steps. 
      Inspired by CopyNet \cite{CopyNet}, the first step uses Generate-Mode to predict a relation. 
      Then, the model switches to Copy-Mode and selects the head and tail entities one by one in two different time steps.
      
      
      \subsection{Encoder}
      
      To model the semantics of the input sentence better, CopyRE adopts Bidirectional LSTM (BiLSTM) \cite{bilstm} as the encoder, which has shown great strength in many areas of NLP. Given a sentence of word embeddings $\{\bm{e}^E_1,...,\bm{e}^E_n\}$ as input, the hidden states from two directions are computed:

      \begin{equation}
      \begin{split}
      \overrightarrow{\bm{h}_i} &= \overrightarrow{LSTM^E}(\bm{e}^E_i, \bm{h}_{i-1}) \\
      \overleftarrow{\bm{h}_i} &= \overleftarrow{LSTM^E}(\bm{e}^E_i, \bm{h}_{i+1})\\
      \bm{h}^E_i &= (\overrightarrow{\bm{h}_i}+\overleftarrow{\bm{h}_i})/2
      \label{eq:encoder}
      \end{split}
      \end{equation}

      where hidden states $\overrightarrow{\bm{h}_i}$ and $\overleftarrow{\bm{h}_i}$ from two directions are averaged\footnote{The original paper uses concatenation, but actually they use average in the released code.} into one vector $\bm{h}^E_i$ as output.

      \subsection{Decoder}
      \label{copyre_decoder}
      
      The decoder uses a one direction LSTM to predict the outputs from left to right. 
      The last hidden state of the encoder is used to initialize the decoder hidden state. 
      The attention score is assigned to each hidden state of the encoder and then summed up to obtain an attentive sum.
      Then the sum is combined with the decoded hidden states at the last time step to be fed into the decoder LSTM:

      \begin{equation}
      \begin{split}
      \bm{c}_t &= Attention(\bm{h}^D_{t-1}, \bm{h}_{1:n}^E) \\
      \bm{u}_t &= [\bm{e}^D_t; \bm{c}_t] \cdot \bm{W}^u \\  
     \bm{h}^D_{t} &= LSTM^D(\bm{u}_{t}, \bm{h}^D_{t-1}) 
      \end{split}
      \label{eq:decoder}
      \end{equation}
      where $Attention$ calculates the attentive sum of all encoder hidden states $\bm{h}_{1:n}^E = \{\bm{h}^E_1,...\bm{h}^E_n\}$ according to the last decoder hidden state $\bm{h}^D_{t-1}$. $[\cdot;\cdot]$ is the concatenation operator, $\bm{e}_t$ is the embedding of  the decoder output in the last time step, $\bm{W}^u \in \mathbb{R}^{(d_e+d_c) \times d_e}$ is the parameter of linear transformation. All biases are omitted for convenience. 
      
      Every three time steps form a loop in which the decoder predicts relation, last token of head and then last token of tail to form a triplet, respectively. The confidence $q^i_{t}$ for each token at position $i$ to be copied as an entity is calculated by:
  
      \begin{equation}
       q^i_{t} = [\bm{h}_t^D;\bm{h}_i^E] \cdot \bm{W}^e \label{eq:problem}
      \end{equation}
      where $\bm{W}^{e} \in \mathbb{R}^{2d_o \times 1}$.
  

      Then, the decoder computes the logits according to the time step $t$ (we count the time step from $1$):

      \begin{equation}
      \bm{logit}_t = \begin{cases}
      [\bm{h}^D_t \cdot \bm{W}^r;q^{NA}],& \text{if} \ t\%3=1; \\
      [\bm{q}_t;q^{NA}],& \text{if} \ t\%3=2; \\
      [\bm{M} \otimes \bm{q}_t;q^{NA}],& \text{if} \ t\%3=0.
      \end{cases}
      \label{eq:mask}
      \end{equation}
      where $\bm{W}^{r} \in \mathbb{R}^{d_o \times rel}$, $rel$ is the cardinality of relations, $\bm{q}_t$ is the concatenation of all $\bm{q}_t^i$, $\bm{M}$ is the mask which records the predicted head entity and prevents the decoder predicting it at the $t\%3=0$ time step. This is based on the fact that an entity cannot be both the head and the tail in the same triplet at the same time. But the mask makes no contributions for minimizing the cross entropy loss we will describe below.

      Through the unnormalized logit, we can obtain the probability of output entity or relation by softmax:
      
      \begin{equation}
      p(y_t|y_{<t}, s) = \frac{e^{logit^j_t}}{\sum_i e^{logit^i_t}}
      \end{equation}
      At time step $t\%3=1$ when the model should predict relation, the softmax score is calculated over all relations types; when the model should predict the entity, the softmax score is calculated over all positions in the source sentence.
      Then, the model can be trained via minimizing the cross entropy loss, which measures the difference between the output $y_t$ and the label $y_t^*$.
      
      \begin{equation}
      \mathcal{L}^D = - \sum_t log(p(y_t^*|y_{<t}, s))
      \label{eq:lossd}
      \end{equation}
      CopyRE also use padding triplets \textit{(NA, NA, NA)} during training,
      which do not have any valid relations and entities. The confidence $q^{NA}$ of NA-relation and NA-position of the corresponding entity is calculated through a shared parameter:
      \begin{align}
      q^{NA} = \bm{h}^D_t \cdot \bm{W}^{NA} \label{eq:na}
      \end{align}
      where $\bm{W}^{NA} \in \mathbb{R}^{d_o \times 1}$.
  
      
      \subsection{Problems of CopyRE}
      \label{sec:problem}
      
      As mentioned in the introduction, we found that CopyRE has two problems.    
      First, the prediction of the entity is unstable. 
      In detailed experiments, we observed that CopyRE cannot even fit the training set well, in which the F1 scores are approximately {0.75} and {0.40} on two datasets (see Fig. \ref{fig:training_curve}). 
      In addition, if we remove the mask $\bm{M}$ in Eq. \eqref{eq:mask}, the F1 score will turn to zero immediately. 
      To find out the reason behind it, we evaluate CopyRE for the predicted relations and entities in the triplets separately.
      The experiments show that CopyRE can gain {0.84} F1 score for relations, while the F1 score for entities dramatically drops to {0.64} (see Table \ref{tab:subtask}). 
      In addition, when we inspected the prediction errors, we find that CopyRE is prone to mix up the order of head and tail.
      Thus, we can conclude that entity copying is the bottleneck of the model, which causes the performance decline.
      
      Second, since CopyRE only predicts the last token of the entity, when the target entity contains multiple tokens, the outputs are incomplete. 
      There are straightforward ways to solve this problem. 
      For example, we can extend the predicted triplets to quintuple by adding the length of entities.
      However, such methods indirectly use or simply ignore the interactions between relation extraction and entity recognition. 
      We propose a multi-task manner method to solve this problem and give detailed comparisons in experiments.

  \section{Our Method}

  As described in the last section, CopyRE is suffering the entity copying and the multi-token entity problems. 
  We propose a model named CopyMTL (Fig. \ref{fig:copymt}) to address these two problems.
  CopyMTL is based on a new model structure and uses a multi-task framework which adds a sequence labeling task to CopyRE encoder.
  In this section, we first reveal the reasons behind the entity copying problem, then propose a simple but reasonable solution.
  After that, we introduce an additional tagging layer of the encoder and the multi-task training procedure.

  \subsection{New Structure for Entity Copying}
  Strangely, in CopyRE entity copying highly depends on the entity mask $\bm{M}$, and the predicted distributions of head and tail entities are identical. 
  Through our analysis, the main culprit is found in Eq. \eqref{eq:problem}, who calculates the concatenation of  $\bm{h}^D_t$ and $\bm{h}^E_i$, then passes it to a linear transformation.
  Eq. \eqref{eq:problem} can be expanded and simplified to get the following form: 
  \begin{figure}[t]
    \centering
    \includegraphics[width=0.4\textwidth]{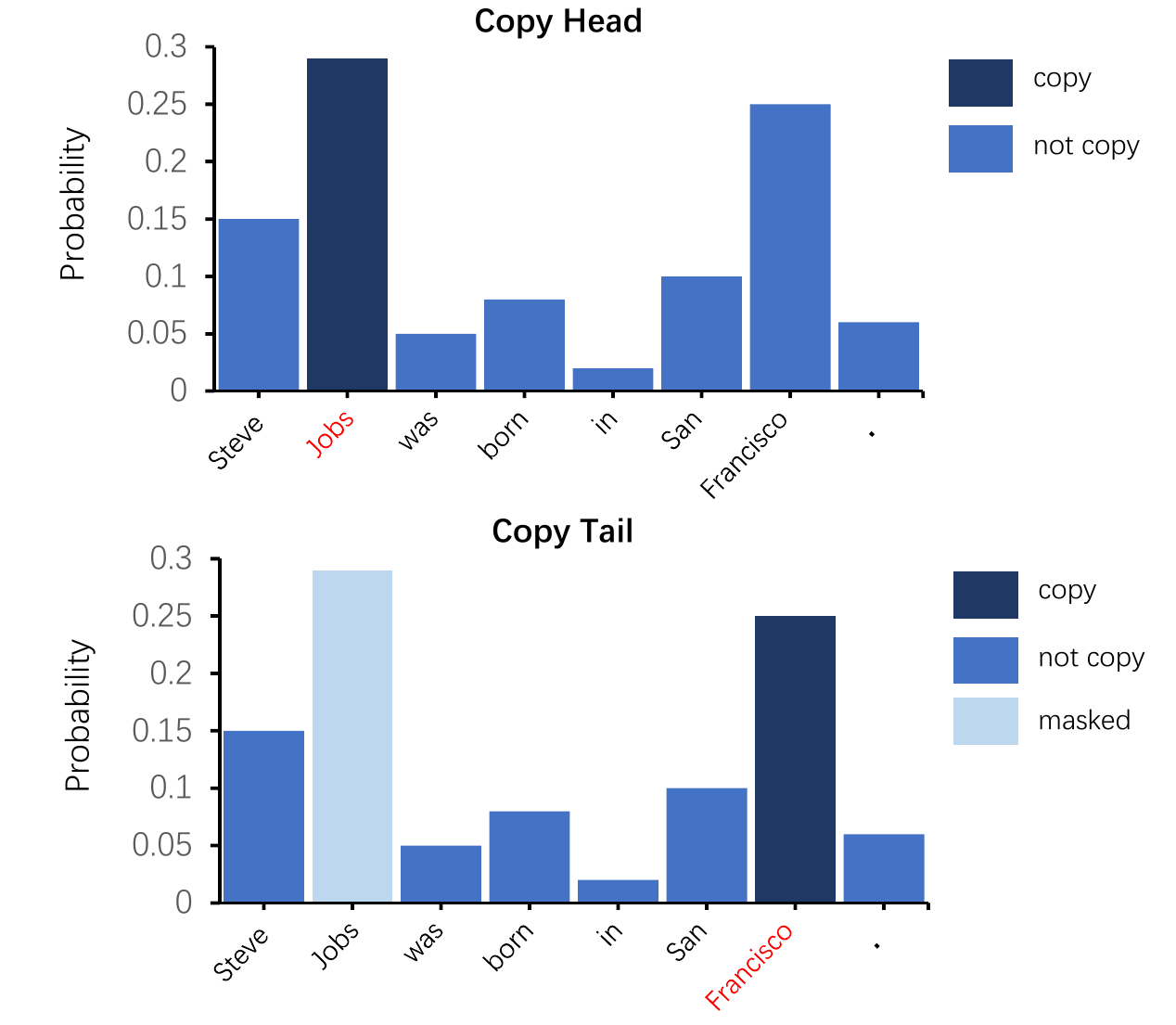}
    \caption{The problematic entity copying of CopyRE. After predicting relation \textit{BirthPlace}, the model will copy the head entity \textit{Jobs}, then mask the predicted head and copy the tail \textit{Francisco}.}
  \label{fig:problem}
  \end{figure}

  \begin{equation}
  \begin{split}
      q^t_{i} &= [\bm{h}_t^D;\bm{h}_i^E] \cdot \bm{W}^e \\
       &= [\bm{h}_t^D;\bm{h}_i^E] \cdot [\bm{W}^e_1;\bm{W}^e_2]\\ 
       &= \bm{h}_t^D \cdot \bm{W}^e_1 + \bm{h}_i^E \cdot \bm{W}^e_2 
  \end{split}
  \end{equation}
  where $\bm{W}^e_1,\bm{W}^e_2 \in \mathbb{R}^{d_o \times 1}$. 
  Note that this is a summation of two scalars and the first term is independent of $i$. 
  If we omit the $q^{NA}$, the probability of entity copying is calculated by softmax:
  
  \begin{equation}
  \begin{split}
      p(y_t|y_{<t}, s)
      &= \frac{e^{q^t_i}}{\sum_j e^{q^t_j}} 
      = \frac{e^{\bm{h}_t^D \cdot \bm{W}^e_1} \cdot e^{\bm{h}_i^E \cdot \bm{W}^e_2}}{e^{ \bm{h}_t^D \cdot \bm{W}^e_1} \cdot \sum_j e^{\bm{h}_j^E \cdot \bm{W}^e_2}} \\
      &= \frac{e^{\bm{h}_i^E \cdot \bm{W}^e_2}}{\sum_j e^{\bm{h}_j^E \cdot \bm{W}^e_2}} \label{eq:reduction}
  \end{split}
  \end{equation}

      \subsubsection{Abnormal dependency to the mask:}In Eq. \eqref{eq:reduction}, we can see that $prob^t_i$ does not rely on the time step $t$. 
      In other words, the output distribution of entity copying at $t\%3=1$ and $t\%3=2$ are identical, which causes the dependency on the mask. 
      We visualize the output distribution of the entity copying in Fig. \ref{fig:problem}. 
      In the figure, the model first copies the token with the highest probability, \textit{Jobs}. 
      Then, in the next time step, as the pointed token \textit{Jobs} is masked, the model copies the token with the second highest probability, \textit{Francisco}.
      
      \subsubsection{Unstable entity copying:}Because the distributions of two time steps are the same and the mask is only used in evaluation rather than optimization, 
      the entity copying, especially for the head entity, becomes unstable. 
      In the training stage, CopyRE maximizes the likelihood for the head at $t\%3=2$ and for the tail at $t\%3=0$, while the likelihood at each time step is identical. However, as the mask is not used for optimization, there is no explicit constraint to ensure that the head has the highest probability and tail has the second highest probability. In fact, CopyRE tries to maximize both the head and the tail. Thus, which one would be the highest and be predicted at $t\%3=2$ is random.

  To fix the problem in Eq. \eqref{eq:reduction}, we simply map $\bm{h}^D_t$ and $\bm{h}^E_i$ to a fused feature space via one additional non-linear layer:
  
  \begin{equation}
      q^t_{i} = \sigma([\bm{h}_t^D;\bm{h}_i^E] \cdot \bm{W}^f) \cdot \bm{W}^o 
      \label{eq:fix}
  \end{equation}
  where $\sigma$ is the $selu(\cdot)$ activation function \cite{selu}, $\bm{W}^f \in \mathbb{R}^{2d_o \times d_{\bm{W}^f}}$ and $\bm{W}^o \in \mathbb{R}^{d_{\bm{W}^f} \times 1}$. 
  
  Due to the non-linearity of the activation function, the reduction of Eq. \eqref{eq:reduction} does not hold true. 
  Now, the entity copying depends on both $i$ and $t$ and there is only one target output to maximize instead of that in Fig. \ref{fig:problem}. 
  Thus, by replacing Eq. \eqref{eq:problem} with Eq. \eqref{eq:fix}, the decoder no longer needs to struggle with ranking head and tail at $t\%3=2$, and the mask is no longer urgently needed\footnote{In experiments, we found that adding the mask to our method brings no enhancement.}. 
  Therefore, the entity copying becomes stable with our new structure. 

\begin{table*}[t]
  \centering
  \begin{tabular}{cccc|ccc}
  \hline
  \multicolumn{1}{c|}{\multirow{2}{*}{Model}} & \multicolumn{3}{c|}{NYT} & \multicolumn{3}{c}{WebNLG} \\ \cline{2-7} 
  \multicolumn{1}{c|}{} & Prec & Rec & F1 & \multicolumn{1}{l}{Prec} & \multicolumn{1}{l}{Rec} & \multicolumn{1}{l}{F1} \\ \hline
  \multicolumn{1}{l|}{NovelTagging} & .642 & .317 & .420 & .525 & .193 & .283 \\
  \multicolumn{1}{l|}{CopyRE-One (ours)} & .612 & .530 & .571 & .312 & .272 & .291 \\
  \multicolumn{1}{l|}{CopyRE-Mul (ours)} & .610 & .566 & .587 & .319 & .273 & .294 \\

  \multicolumn{1}{l|}{GraphRel-1p} & .629 & .573 & .600 & .423 & .392 & .407 \\
  \multicolumn{1}{l|}{GraphRel-2p} & .639 & .600 & .619 & .447 & .411 & .429 \\
  \multicolumn{1}{l|}{CopyMTL-One} & .727 & \textbf{.692} & .709 & {.578} & \textbf{.601} & \textbf{.589} \\
  \multicolumn{1}{l|}{CopyMTL-Mul} & \textbf{.757} & {.687} & \textbf{.720} & \textbf{.580} & .549 & .564 \\ \hline
  \end{tabular}
  \caption{Results of the compared models on NYT and WebNLG, in which CopyRE uses less strict evaluation.}
  \label{tab:main_result}
  
  \end{table*}

  \begin{table}[t]
    \centering
    \begin{tabular}{lcc}
    \toprule
    Dataset & NYT &  WebNLG \\
    \midrule
    Relation types & 24 &  246 \\
    Dictionary size & 90760 &  5928 \\
    Train sentence & 56195 & 5019 \\
    Test sentence & 5000 &  703 \\
    \bottomrule
    \end{tabular}
    \caption{Statistics of the datasets. }
    \label{tab:statistics}
    \end{table}
  \subsection{Sequence Labeling Layer}
  
  CopyRE only copies the last token of the entity. 
  To predict entities with multiple tokens, we cast the problem into a sequence labeling problem and use the NER results to calibrate the entities with multiple tokens. 
  As shown in Fig. \ref{fig:copymt}, we first derive the emission potential from the encoder output.
  Then, an additional Conditional Random Field (CRF) layer \cite{crf} is employed to calculate the most probable tag for each token.
  We use the BIO scheme (Begin, Inside, Outside) to recognize all of the entities in the sentence. 
  The predicted tags are used to post-process the extracted entities. 
  
  The conditional probability of target $tags^*$ given sentence $s$ are computed by path probability:
  
  \begin{equation}
      p(tags^*|s) = \frac{e^{score(s, tags^*)}}{\sum_{tags'}e^{score(s, tags')}}
  \end{equation}
  where the denominator is computed via dynamic programming. The unnormalized path score is defined as: 
  
  \begin{equation}
      score(s,tags) = \sum_{i} \phi_{i,tag_i} + b_{tag_{i-1}\rightarrow tag_i}
  \end{equation}
  where $b_{tag_{i-1}\rightarrow tag_i}$ is the transition score from $tag_{i-1}$ to $tag_i$.
  $\phi_{i,tag_i}$ is the score of the $tag_i$ for the $i$-th input token, which is comes from the hidden state of the Bi-LSTM at timestep $i$.
  
  The loss function of the sequence labeling is:
  
  \begin{equation}
      \mathcal{L}^E = -log(p(tag^*|s))
  \label{eq:losse}
  \end{equation}
  
  In the inference stage, we use the NER results to post-process the decoded entities. Since we use BIO tagging scheme, there are three circumstances for the decoded last token of entities:
  \begin{itemize}
    \item 'B'. a single token entity.
    \item 'I', an entity with multiple tokens, it will look for the token before the current token until it finds 'B'. 
    \item 'O', a single token entity.
  \end{itemize}


  \subsection{Training}
  
      Overall, the input sentence is fed into the encoder part. 
      All of the hidden states of the encoder are used to label the input sequence and calculate the attention of the decoder. 
      Initialized by the last hidden state of the encoder, the decoder generates triplets each three time steps. 
      Thus, the loss function contains two parts: the encoder part introduces an additional CRF loss, and in the decoder part the cross entropy loss is used to measure the difference between the output triplets and the gold triplets. 
  
      We define the loss function as the weighted summation of encoder loss and decoder loss:
  
  \begin{equation}
      \mathcal{L}= \lambda \cdot \mathcal{L}^E +\mathcal{L}^D
  \end{equation}
  where $\lambda$ is the weight of the tagging loss.
  
  The loss is calculated as the average over shuffled mini-batch, and the derivatives of each parameter can be computed via back-propagation. 

  \section{Experiments}



  \subsection{Datasets and Setting}

      We evaluated models on two datasets: New York Times (NYT) \cite{nyt} 
      and WebNLG \cite{webnlg}. 
      NYT comes from the distant supervised relation extraction task (DSRE), which aims to leverage the strength of the knowledge base to generate a large-scale dataset \cite{dsre}. To make joint extraction more challenging than DSRE experiment setting, \citeauthor{CopyRE} \shortcite{CopyRE} additionally modified the data to include more overlapping relations.
      WebNLG is originally used for natural language generation, in which all of the sentences are written by annotators. To avoid that the model only remembers the entity linking instead of the relation pattern, we only use the first sentence for each instance, which is the same as other baselines. 
      The data statistics of both datasets are shown in Table \ref{tab:statistics}.  

      Our experiments settings also followed most of the settings of CopyRE.
      The hidden number of LSTM was set to 1000.  
      The max number of decoded triplets was 5. This was because the average triplet number in both dataset is about 2. We did not use ``end-of-sentence'' token to stop decoding, but to decode all padding triples \textit{(NA, NA, NA)}.
      The embedding dimension was 100, and we used the same pretrained embeddings\footnote{\url{https://github.com/xiangrongzeng/copy_re}}.
      Adam \cite{Adam} was used to optimize the neural networks and the learning rate was 0.001. The weight of $L^E$, $\lambda$, was set to 1. 

  \subsection{Baselines and Evaluation Metrics}
   

      We compare CopyMTL with CopyRE \cite{CopyRE}, NovelTagging \cite{novel_tagging} and GraphRel \cite{graphrel}. NovelTagging uses sequence labeling to assign one label to each word, which contains both entity and relation information. GraphRel is the state-of-the-art model, which uses a post-editing method to revise the triplets phase by phase. For Seq2Seq model, CopyRE and our CopyMTL, we give a more detailed comparison to show the advantages of our new structure.
      We also evaluate the OneDecoder and MultiDecoder trick for the Seq2Seq models (denoted as -One and -Mul).
      The main difference between the two decoders is the parameter sharing strategy. OneDecoder uses shared parameters for predicting all triplets and is exactly what we described in the background section.
      MultiDecoder uses unshared decoders, each decoder predicts one triplet.

      We use precision, recall, and micro-F1 score to evaluate the models. 
      The evaluation metrics we use are stricter than that of the original CopyRE.
      That is to say, instead of leaving out the incomplete entity problem, the outputs of our experiments are regarded as correct only if both the relation types and all entity tokens are correct.     
      This stricter metric meets real-world usage and the comparison is fairer to NovelTagging and GraphRel because they are not haunted by the multi-token problem. For an intuitive comparison, we also list the result of CopyRE in the table, although their evaluation is not so strict.
      \begin{figure*}[t]
        \centering
        \resizebox{0.75\textwidth}{!}{
          \subfigure[]{\label{eg:a}
          \begin{minipage}[t]{0.5\linewidth}
            \centering
          \includegraphics[width=1.0\textwidth]{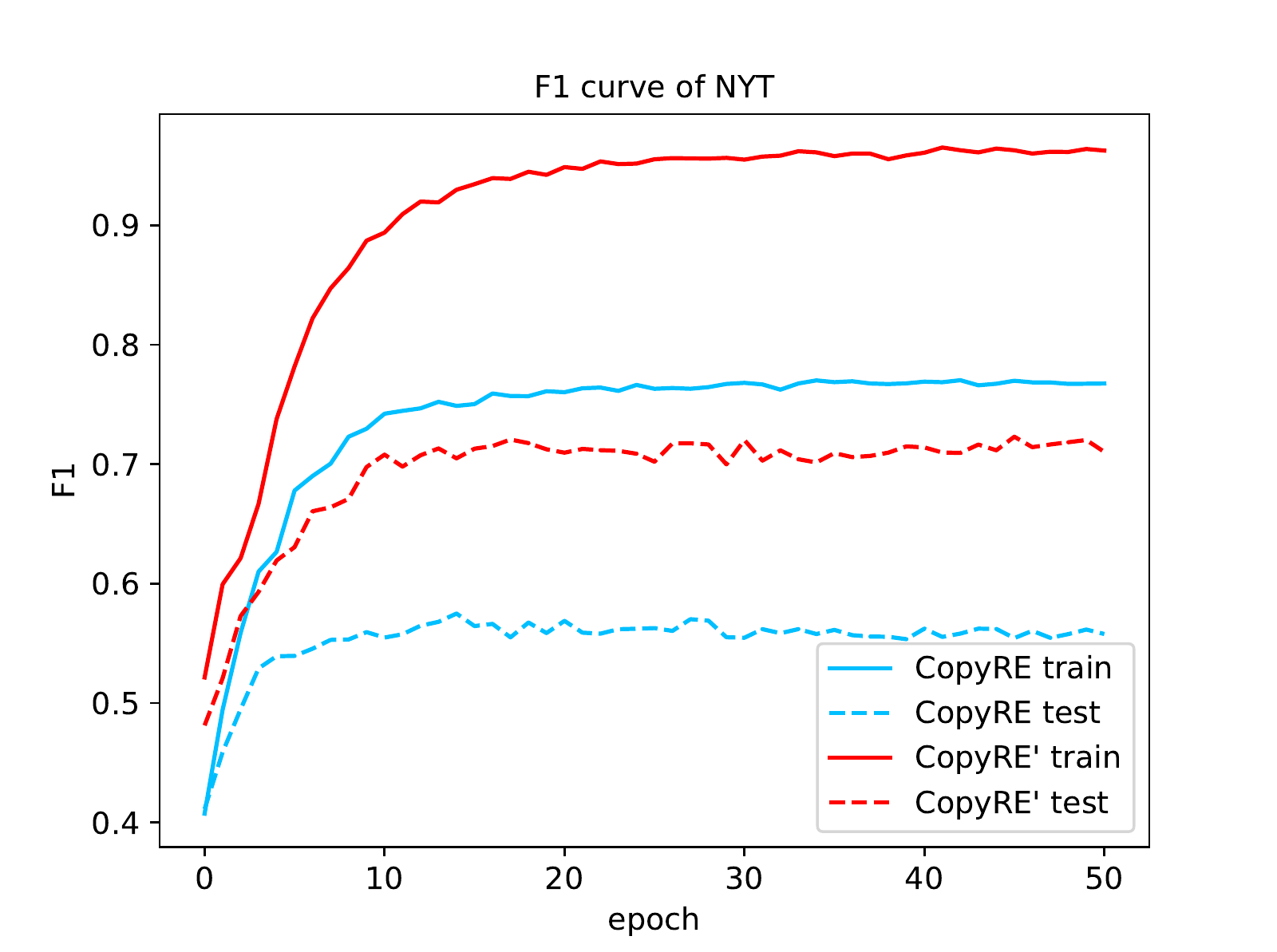}
          \end{minipage}
        }
          \subfigure[]{\label{eg:b}
          \begin{minipage}[t]{0.5\linewidth}
            \centering
          \includegraphics[width=1.0\textwidth]{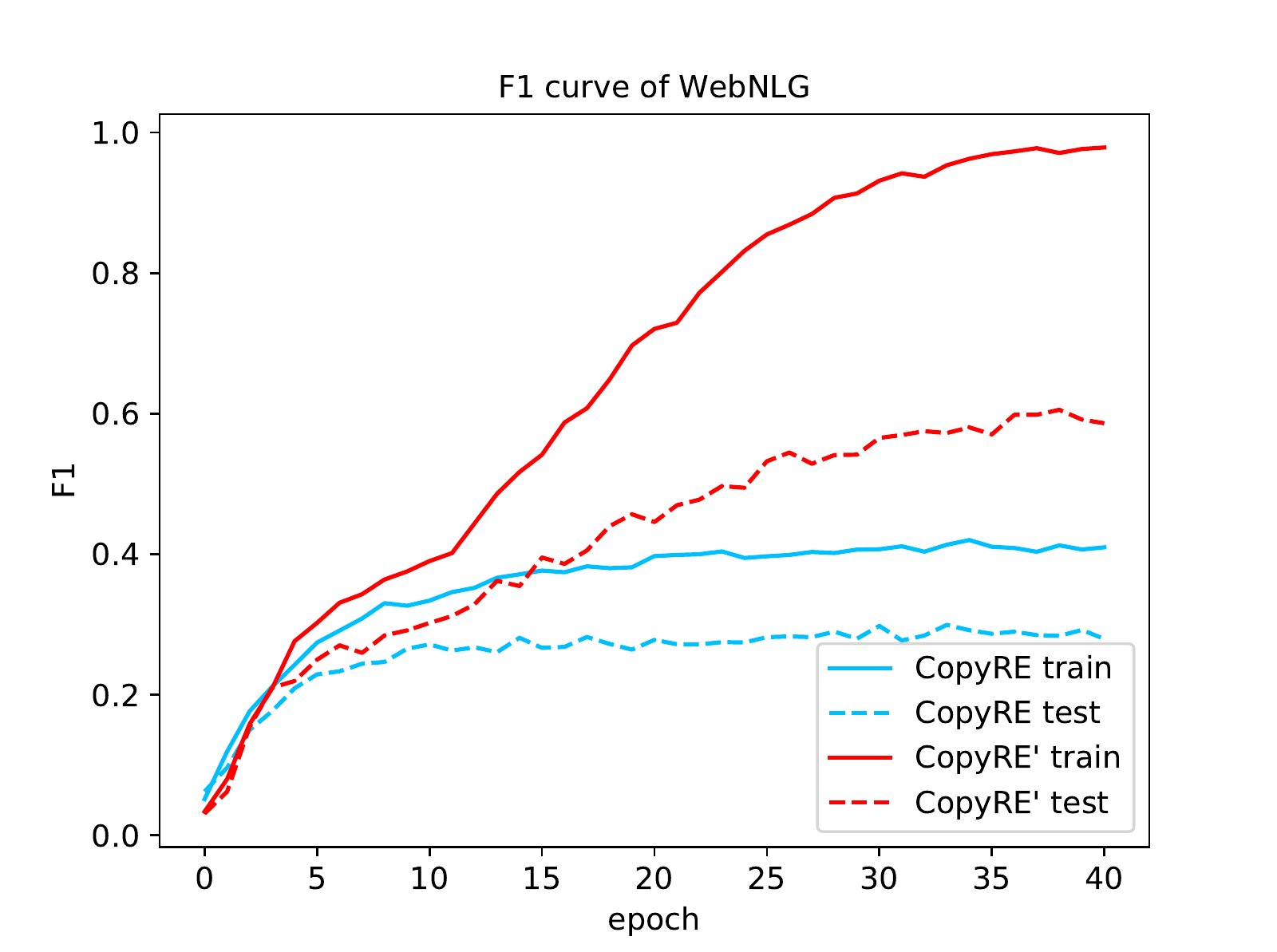}
          \end{minipage}
        }
        }
        \caption{The training curves of CopyRE and CopyRE' on NYT and WebNLG. }
          \label{fig:training_curve}
        \end{figure*}
  
  \subsection{Comparison of Baselines}

      To evaluate the performance of the proposed method, we compare CopyMTL with the baseline methods.
      The results\footnote{As NovelTagging is significantly better than previous works, we do not add more comparisons.} are shown in Table \ref{tab:main_result}.
      
      As shown, CopyMTL is the best model in WebNLG and NYT.
      Both the precision and the recall are significantly improved. 
      In the NYT dataset, compared with the state-of-the-art model, {GraphRel-2p}, CopyMTL-One outperforms it by 8.8\% for precision and 9.2\% for recall.     
      In the WebNLG dataset, the effect is more significant.
      The improvements are 13.1\% in precision and 19\% in the recall.
      These observations verify the effectiveness of our proposed method.
      NovelTagging is characterized by a low recall, which is caused by its deficiency in overlapping relations. CopyRE has already solved this problem well, with 8\% and 19\% absolute F1 improvement in WebNLG and NYT. Our method further brings 33\% and 19\% F1 enhancement compared to CopyRE, which shows great potentials of Seq2Seq methods. 
    
      CopyRE argued that MultiDecoder is better than OneDecoder, which is validated by our reproduction experiment. However, with our novel CopyMTL, MultiDecoder is better than OneDecoder in NYT but worse in WebNLG. This is probably because NYT is a bigger dataset, in which MultiDecoder with more parameters works better. In practice, which decoder to use should be determined by the size of the dataset and we cannot conclude that one is better than another in every situation. For simplicity, we only discuss OneDecoder in the following sections.
  

  \subsection{Effects of the Revised Entity Copying Method}
  \label{sec:curve}
  
  \begin{table}[t]
  \centering
  \begin{tabular}{l|l|ccc}
  \hline
  Dataset & Model & \multicolumn{1}{l}{Prec} & Rec & F1 \\ \hline
  \multirow{2}{*}{NYT} & CopyRE & .612 & .530 & .571 \\
   & CopyRE' & \textbf{.747} & \textbf{.700} & \textbf{.722} \\ \hline
  \multirow{2}{*}{WebNLG} & CopyRE & .312 & .272 & .291 \\
   & CopyRE' & \textbf{.583} & \textbf{.629} & \textbf{.605} \\ \hline
  \end{tabular}
  \caption{Results of CopyRE and CopyRE' on NYT and WebNLG. These models do not consider entities with multiple tokens and use less strict evaluation that ignores entity with multiple tokens.}
  \label{tab:first_problem}
  \end{table}

      Although CopyMTL outperforms baselines by a huge margin, it is still unclear which component in CopyMTL plays the pivot role. 
      To reveal the strength of the new model architecture, we compare CopyRE with the modified model, called CopyRE', which only substitutes Eq. \eqref{eq:problem} for Eq. \eqref{eq:fix}. 
      The comparison is in Table \ref{tab:first_problem}, from which we can observe that Eq. \eqref{eq:fix} is extremely effective. 
      CopyRE' model gains 13\% F1 boost in NYT dataset and 31\% F1 boost in WebNLG dataset.

      Note that the new model architecture only considers the entity copying while the F1 score computed considers the whole triplet. 
      In order to uncover the performance of CopyRE' in relation classification and entity recognition, we calculate the F1 scores for the two subtasks in Table \ref{tab:subtask}.  
      For the entity recognition subtask, the F1 score of CopyRE' is 10\%  higher in NYT and 19\% higher in WebNLG. This is the main contribution of the new model architecture. 
      For the relation classification subtask, the F1 score of CopyRE' is marginally higher (less than 3\%) than that of CopyRE. 
      This implies the better entity recognition helps relation classification learning, which confirms the argument that the interactions between two task are beneficial to each other. 
      In the decoding stage, a more precise prediction of the entity is fed into the decoder, which aids the relation classification in the next time step.
      
      \begin{table}[t]
          \centering
          \begin{tabular}{l|l|cc}
              \hline
              Dataset & \multicolumn{1}{l|}{Model} & Relation & \multicolumn{1}{r}{Entity} \\ \hline
              \multirow{2}{*}{NYT} & CopyRE & .846 & .647 \\
              & CopyRE' & \textbf{.869} & \textbf{.756} \\ \hline
              \multirow{2}{*}{WebNLG} & CopyRE & .767 & .595 \\
              & CopyRE' & \textbf{.797} & \textbf{.782} \\ \hline
          \end{tabular}
          \caption{F1 scores on subtasks.}
          \label{tab:subtask}
      \end{table}
      


      Except for the final result, the learning processes of the two models are also different. 
      We plot the overall F1 score varying with the training epochs in Fig. \ref{fig:training_curve}. 
      The curve shows that CopyRE does not fit the training set well and the model saturates at epoch 20, where the F1 score of NYT is 75\% and the F1 score of WebNLG is 40\%. 
      By contrast, CopyRE' gains 97\% F1 score in the NYT training set and 97\% F1 score in the WebNLG training set. 
      In addition, the performance of CopyRE' continues increasing until epoch 40 on both datasets. 
      The fact that the model gains lower training error which also generalizes to the test set may explain the effectiveness of CopyRE'.

  

  \subsection{Effects of Multi-Task Learning}
  \label{sec:quintuple}
  
      CopyMTL aims to solve the multi-token problem. 
      In addition to the multi-task learning used by CopyMTL, there can be other straightforward methods. 
      For example, 
      the decoder of CopyRE' can predict the length of the entity when it copies the entities, which forms quintuples, called CopyRE'5. This is similar to predicting both the begin and the end of the entities and should work the same.
  
      We compare the models in Table \ref{tab:copy5}, from which we can see that CopyRE'5 is worse than CopyMTL in all evaluations, but both outperform GraphRel. 
      We conjecture that the three tasks in CopyRE'5, include relation classification, entity recognition, and entity length prediction, varying in their degree of difficulty.
      The entity length prediction task may interfere with the learning of other tasks, as this easier task prolongs the dependence distance of harder tasks. 
  
      In addition, we also evaluate how precisely does the encoder of CopyMTL completes the whole entities. It gains 99\% F1 score in NYT and 96\% F1 score in WebNLG. This evaluation is less strict than conventional NER tasks as we consider neither the types of the entities nor the entities out of relations. We can conclude that NER in joint extraction is powerful enough for triplet extraction and the main difficulty in joint extraction is to make a better prediction for both the relations and the positions of the corresponding entities.

      \begin{table}[t]
        \centering
        \begin{tabular}{l|l|ccc}
        \hline
        Dataset                 & Model       & Prec          & Rec           & F1            \\ \hline
        \multirow{3}{*}{NYT}    & GraphRel-2p & .639          & .600          & .619          \\
                                & CopyRE'5    & .680          & .663          & .671          \\
                                & CopyMTL     & \textbf{.727} & \textbf{.692} & \textbf{.709} \\ \hline
        \multirow{3}{*}{WebNLG} & GraphRel-2p & .447          & .411          & .429          \\
                                & CopyRE'5    & .572          & .536          & .553          \\
                                & CopyMTL     & \textbf{.578} & \textbf{.601} & \textbf{.589} \\ \hline
        \end{tabular}
        \caption{Results of different multi-token models on NYT and WebNLG}\label{tab:copy5}

        \end{table}

  \section{Related Work}

  Extraction of entities and relations is of significance to many NLP tasks. In recent years, there have been four mainstream methods.
  
   \textbf{Pipeline methods}: Previous works mainly use pipeline methods \cite{pipeline1}, a.k.a extract entities first then classify the relations. 
  Most of the recent neural models also focus on pipeline methods, include (1) Fully-Supervised Relation Classification \cite{re_full} (2) Distant Supervised Relation Extraction \cite{dsre}. In spite of the recent progress of neural models  \cite{nrc_lstm,zeng14,nrc_walk,noise3},  the pipeline methods introduce error propagation problem \cite{interaction}, which does harm to the overall performance.

   \textbf{Table filling}: The joint extraction task is formalized as a table constituted by the Cartesian product of the input sentence to itself. 
      The table blanks, except for that on the diagonal, are to be predicted as relations.
      The models include history-based searching \cite{14table_fill}, neural-based prediction \cite{table_filling} and global normalization \cite{crf_norm}. The state-of-the-art model, GraphRel, also belongs to this genre. This model innovative takes the interaction between entities and relations into account via 2-phase GCN. The main problem of table filling is the over redundant computation for the permutation of all word pairs in a sentence. As a result, most of the blanks in the table are empty and it is the sparsity that hinders the learning of the models.
  
  
   \textbf{Tagging}: The tagging models originally solved the tasks separately through a shared parameter: the model tags the entities first, then predicts the relations. 
      SPTree \cite{tree} used a structural neural model with the help of linguistic features. 
      This model was promoted by an attention-based model \cite{limb}. 
      Besides, NovelTagging \cite{novel_tagging} proposed a new tagging scheme, by which the model can predict a single tag for each word, containing both the entities and relations. However, this tagging scheme cannot handle overlapping relations because it cannot assign one token with multiple labels. 
  To solve it, multi-pass tagging training, HRL\footnote{We did not use it as a compared baseline because HRL requires complicated preprocessing procedure and is not transferable to different datasets. 
  }, has been purposed \cite{rlre}, based on the reinforcement learning framework and \cite{tag2table} leverage attention mechanism. Although these methods solve the overlapping relation problem, their nature and complexities are akin to table filling.
  
   \textbf{Seq2Seq}: CopyRE \cite{CopyRE} is another method for solving the overlapping relation problem, which extracts triplets by a Seq2Seq framework \cite{Seq2Seq} with copy mechanism \cite{CopyNet}. But it cannot predict the entire entities. In addition, the weak performance hinders it from real-world usage. 
   Our work resolves the problems and boosts the performance to a new level.
  
  \section{Conclusions and Future Work}
  
  In this paper, we revisit the CopyRE model, which jointly extracts entities and relations by a Seq2Seq model. We find that there are two problems in the model: the performance of the model is limited by the inaccurate entity copying and the generated entities are not complete. We give a theoretical analysis to reveal the reason behind the first problem, then propose a new model architecture to solve it. For the second problem, we propose a multi-task learning framework to complete the entities. Detailed experiments show the effectiveness of our method, which also outperforms the current state-of-the-art model by a huge margin.
  
  For future work, CopyMTL still has much potential, for example, the current model can only extract a fixed number of triplets. We would also like to extend CopyMTL to extract any number of triplets. CopyMTL can build a strong baseline for future studies.

  \section*{Acknowledgment}
  This work was supported by the National Natural Science Foundation of China(No.61602059, 61972057), ``Double First-class'' International Cooperation and Development Scientific Research Project of Changsha University of Science and Technology: 2018IC25. 
\fontsize{9.5pt}{10.5pt} \selectfont
\bibliography{emnlp-ijcnlp-2019}
\bibliographystyle{aaai}
\end{document}